\documentclass[twocolumn,11pt]{article}

\usepackage{listings}
\usepackage{amssymb}
\usepackage{amsmath}
\usepackage{booktabs}
\usepackage{soul}
\usepackage{tikz}
\usepackage{tabularx}
\usepackage[alph,breakwithin]{parnotes}
\usepackage{array,multirow,graphicx}
\usepackage{hyperref}
\usepackage{cleveref}
\usepackage{pifont}
\usepackage[numbers]{natbib}
\usepackage{courier}
\usepackage[margin=0.9in]{geometry}

\newcommand{\cmark}{\ding{51}}%
\newcommand{\xmark}{\;--\;}%
\newcommand{\D}{\mathrm{d}}

\begin{document}

\title{ADSEQ: A delay-aware autograd-compatible framework for
spike-event delivery in SNNs
}

\author{
Lennart P. L. Landsmeer$^{1,2}$\footnotemark[2],
Amirreza Movahedin$^{1,2}$,
Said Hamdioui$^1$,
Christos Strydis$^{2,1,}$\footnotemark[3]
\\
\\
\small
$^1$ Department of Computer Engineering,
Delft University of Technology, Netherlands
\\
\small
$^2$ NeuroComputingLab, Department of Neuroscience,
Erasmus Medical Center, Netherlands
\\
\small
{\footnotemark[2]  \hspace{0.1em} l.p.l.landsmeer@tudelft.nl}
\hspace{3em}
{\footnotemark[3] \hspace{0.1em} c.strydis@erasmusmc.nl}
}

\twocolumn[
  \begin{@twocolumnfalse}
    \maketitle
\begin{abstract}
Efficient simulation and gradient-based training of brain simulations and spiking neural networks (SNNs) require handling sparse spike transmission -- including heterogeneous synaptic delays -- without resorting to dense, memory-intensive representations. Existing, exact-gradient methods are restricted to simplified neuron models and cannot generalize to full-brain simulations or arbitrary queue structures. Moreover, current machine learning (ML)–based simulators either omit delayed spike delivery or rely on ring buffers that scale poorly on modern accelerators.
In this work, we introduce \emph{ADSEQ} (AutoDifferentiable Spike-Event Queues), a generalized, auto-differentiable (AD) framework for spike-event delivery that supports heterogeneous delays, arbitrary queue data structures, and applicability across both artificial and biophysical SNNs. 
To enable delay-gradients in an AD system, we derive the required implicit derivatives and inject these into the AD flow via custom gradients.
We implement a suite of differentiable queue designs --including ring buffers, FIFO structures, binary heaps, sorted arrays, and GPU-optimized variants-- in JAX and benchmark them on CPU, GPU, TPU, and LPU.
Our results show different optimal queue structures given hardware architecture, spike rate, and accuracy constraints. Tree- and FIFO-based queues were found to be optimal on CPU. Ring buffers excel on GPU for smaller networks, while FIFOs scale more favorably for larger networks. TPU benefits from sort-based implementations due to dedicated sparse-core intrinsics, and LPU favors deterministic dataflow scheduling. We further quantify spike-drop rates for lossy queues, demonstrating controllable performance–accuracy trade-offs.
These findings establish ADSEQ as a foundation for scalable, reusable set of components for autograd-compatible artificial and biologically realistic SNN simulations with delays, across AI-accelerator platforms.
\end{abstract}
  \end{@twocolumnfalse}
]

\clearpage

\section{Introduction}

Computational neuroscience and neuro-inspired machine learning (ML) require fast simulation and efficient tuning of artificial and biorealistic spiking neural networks (SNNs). These SNNs consist of local, dense computations in addition to the exchange of sparse spike events between neurons.~\cite{carnevale2006neuron,zhang2020supervised, fernandez2025homeostatic}. Furthermore, to increase the representational power and temporal accuracy of SNNs, in addition to moving towards more biorealistic and accurate modeling, synaptic delays between neurons -- and hence delayed spike-delivery -- has gained increased attention recently~\cite{goltz2024delgrad,meszaros2025efficient}. 

As gradient-free approaches are known to suffer from the curse of dimensionality~\cite{blondel2024elements}, gradient-based optimization methods are extensively used for training SNNs, and increasingly, brain models~\cite{goltz2024delgrad,neftci2019surrogate,muller2024jaxsnn,deistler2024differentiable,landsmeer2024gradient}.
However, gradient calculation for many parameters, now exacerbated by delayed spike delivery and associated delay parameters, makes training increasingly more compute and memory intensive.
To this end, in pursuit of suitable hardware for faster simulation and gradient-based training of SNNs, rapidly developing AI accelerators have been seen as promising substrates~\cite{landsmeer2024gradient,finkbeiner2024harnessing}. SNNs -- both biorealistic and ML-oriented ones -- have been shown to perform very well on novel AI accelerators by using ML libraries --such as TensorFlow or JAX-- to express their computational models~\cite{deistler2024differentiable,landsmeer2024tricking,wang2023differentiable}.

\begin{figure}[t!]
    \centering
    \includegraphics[width=1.0\linewidth]{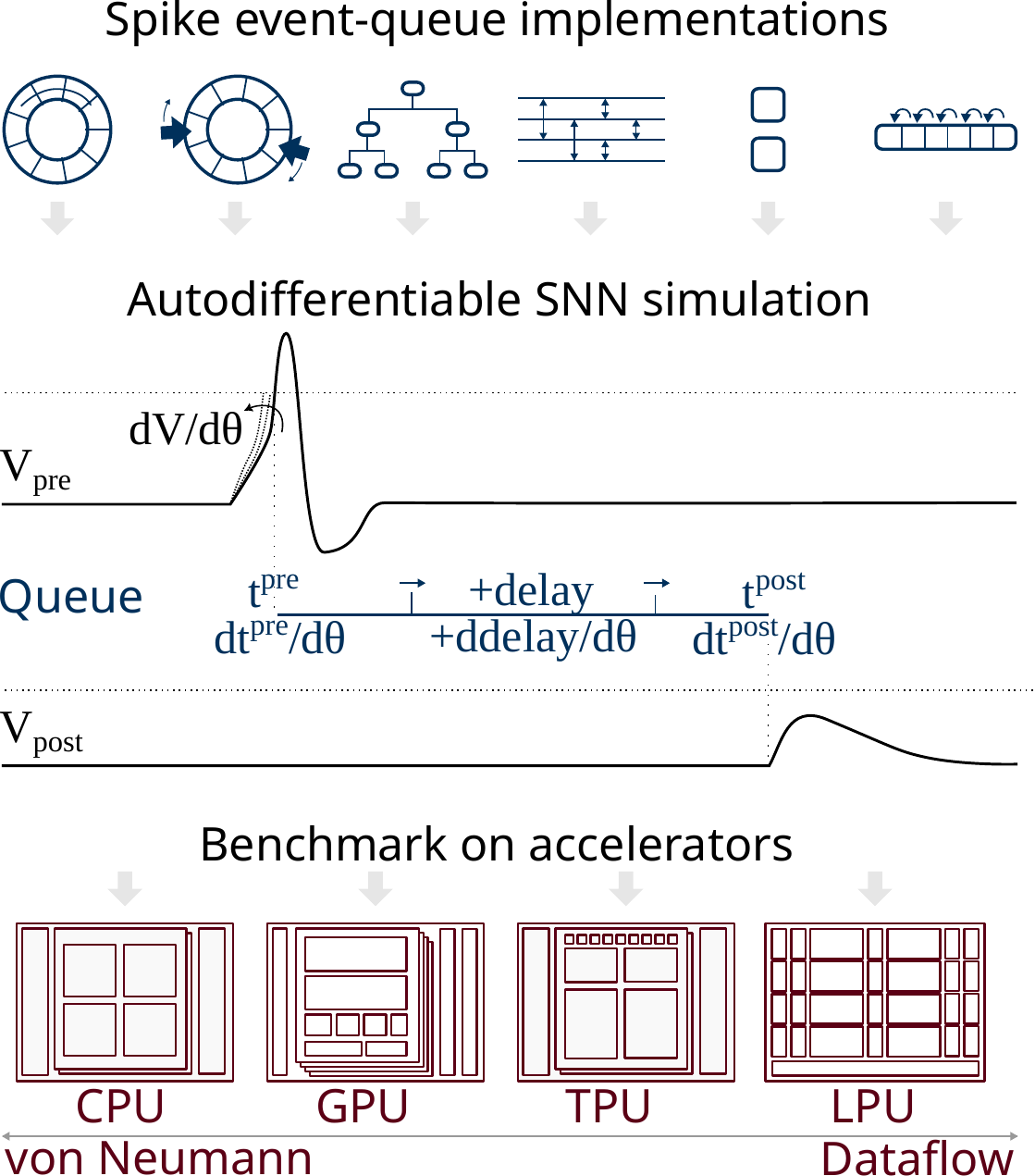}
    \caption{Experimental setup. We implement spike-event queues that support (delay-)gradients and benchmark how various queue structures perform on different AI accelerators.}
    \label{fig:overview}
\end{figure}

Existing gradient-enabled frameworks for SNN training, to overcome problems with the discontinuities resulting from spike-events,
have either employed methods based on surrogate gradients --smoothed gradients of the step function-- or exact gradients --implicit differentiation through discontinuities.
The dominant solution for calculating gradients through discrete spike events has been the former for training SNNs~\cite{neftci2019surrogate,bindsnet,spytorch19,rasmussen2019nengodl,muir_dylan_2019_4639684,SpikingJelly,norse2021,eshraghian2021training,heckel2024spyx}, while realistic brain simulators have evaded the problem by not implementing delayed spike delivery~\cite{deistler2024differentiable}. Although surrogate gradients work well for SNN models without delayed spike delivery, the smoothed surrogate gradient is now most of the time non-zero.
Thereby, this method turns the generally very sparse spike events into a memory-intensive, continuous time signal. This prevents exploiting the sparsity of spike events in time, hence limiting the efficiency of surrogate gradients in the presence of delays.

Exact solutions to spike-event gradients have also been developed~\cite{goltz2024delgrad,muller2024jaxsnn,wunderlich2021event,turner2022mlgenn}. However, while efficient, these only pertain to simplified (i.e. Leaky Integrate-and-Fire, LIF) neurons and have not been shown to be transferable to realistic brain simulations.
The majority of simulators do not support delayed connections and spike delivery at all; the few implementations that do, all use memory-inefficient implementations.
As such, there is a dual need to 1) improve ML library-based SNNs and brain simulations by adding gradient-enabled, delayed spike-event delivery and, 2) to understand how spike delivery can run efficiently across different AI-accelerator architectures.

Delayed spike-event delivery is commonly implemented using heap-based priority queues, ring buffers or First-In First-Out (FIFOs) queues~\cite{wang2023differentiable,arora2009computational,abi2019arbor}.
From these, only ring-buffers allow for gradient calculation, via the aforementioned (memory inefficient) surrogate gradients.
To find the most compute efficient data-structure for each AI-accelerator,
in theory, algorithmic-complexity analysis reveals the most efficient event-queue implementations to be heap-based data structures~\cite{arora2009computational}. This holds under the assumption of standard models of computation,
including the ability to short-circuit loops and diverging/converging conditional control flow.
However, performance differences might manifest across different (more dataflow-oriented) hardware architectures, in practice.
For example, GPUs operate best when employing coalesced memory accesses across parallel execution units and non-diverging code paths, which might shift the balance from heap-based data structures to more simplified data structures in practice. On dataflow architectures, as can be found in many AI accelerators like the Groq LPU~\cite{abts2022groq}, execution is fully deterministic, which leads to the absence of code branching. This results in always having the worst-case time complexity for heap-based algorithms, as all loops need to be fully unrolled. On the other hand, these architectures offer massive parallelization, which might offset the efficiency-loss from worst-case time complexity, as long as workloads fit in memory fully.

In this work, we first develop the methods that allow automatic gradient calculation through arbitrary data structures for both neural parameters and synaptic delays~(\cref{fig:overview}); then, we benchmark the performance of different data structures for delayed spike-events across a variety of AI accelerators that use these methods. We demonstrate that, for event queues, heap-based data structures perform best on von Neuman-style architectures, while sort-based data structures are more fitting for dataflow-style hardware substrates, and structures like ring-buffers are efficient on GPUs as long as the workload fits in cache. Concisely, the contributions of this work are as follows:

\begin{itemize}
    \item An autodifferentiation-compatible formulation of delay-aware exact spike-event gradients.
    \item A collection of autodifferentiable (spike) event-queue data structures expressed in a ML library with custom gradients to inject these extra gradients; ADSEQ.
    \item A cross-accelerator benchmarking study of these autodifferentiable queues, showing architecturally-dependent optimal queues.
    \item A spike dropping mechanism providing a tunable performance–accuracy trade-off.
\end{itemize}

This work is structured as follows:
In \cref{sec:bg}, we discuss the necessary background and go over the related works. In \cref{sec:methods}, the methods for gradient calculations are presented. These gradients are implemented as reusable queue primitives --the ADSEQ library-- in \cref{sec:impl}. For evaluation, the benchmark workloads and platforms are discussed in \cref{sec:eval}. In \cref{sec:res}, the experimental results are presented and these are discussed in \cref{sec:discuss}. Finally, \cref{sec:conc} concludes the work.

\section{Background and related works}
\label{sec:bg}

\subsection{Models of neurons and synapses}

Neuron models vary from simplified to biorealistic. 
Neurons are modeled as connected compartments, each with a different voltage. Simplified neuron models, like  LIF cells, only have a single compartment, and do not use this terminology (also known as a point neuron).
When the voltage in a compartment crosses a threshold, a spike is registered originating from that compartment.
Simplified models usually explicitly model the reset mechanisms after a spike, requiring careful treatment when calculating their gradients.
In biorealistic models, action potentials and resets originate from the action of opening and closing of ion-channel models in a simulated membrane, using stiff ordinary differential equations (ODEs). The prototypical example of this type is the Hodgkin-Huxley model~\cite{hodgkin1952quantitative}. 

Neurons communicate via synapses. A synapse can be \emph{chemical} or \emph{electrical}.
Chemical synapses transfer spikes from a presynaptic, sending neuron, to a postsynaptic, receiving neuron.
Electrical synapses allow bidirectional communication of both spikes and membrane voltage.
Electrical synapses provide instantaneous connection, hence do not require queues when modeling them.
Additionally, as this instant communication via electrical synapses on various AI accelerators have been explored by our previous work~\cite{landsmeer2024tricking}, here we will focus on chemical synapses alone.

When simulating brain models, various simplification on the types of synapses are performed. In traditional brain simulators, all spikes are delivered as separate events on the postsynaptic neuron.
Under the assumption of \emph{linear time invariance (LTI)}, spike events create synaptic responses that can be linearly added together. As such, multiple spikes falling into the same delivery timestep can be summed together, simplifying the delivery mechanisms.
Even further, some simulators decide to trade implementation simplicity for the possibility of dropping spikes~\cite{panagiotou2022eden}. We will refer to this spike-handling optimization as lossy event queues.

Moreover, synaptic delays can be classified as \emph{homogeneous} or \emph{heterogeneous}.
Homogeneous delays have the exact same delay value for all spikes between populations, which removes the requirement to sort the incoming spikes in a priority queue. For the more biorealistic heterogeneous delays, the possibility arises that a slow spike is sent earlier than another faster arriving spike, which requires sorting in the form of a min-heap queue.

In biology, delays in spike transmission result from both the inherent delay of a chemical synapse of around 0.3-0.5ms as well as the physical distance spikes have to travel along the axon and, to a lesser extent, dendrites (the neural processes). When it comes to model accuracy, the latter are usually not interesting to model explicitly at the compartment level comes at high computational costs for just delaying the action potential. Hence, to achieve higher computational efficiency, this type of delay can be replaced by a single ms-order delay value that models both the spike propagation time along the neural processes and the synapse itself.
In AI applications, delays have been shown to enhance the computational power of spiking neural networks, especially w.r.t. to temporal precision~\cite{meszaros2025efficient,landsmeer2024gradient}.
In fact, training of synaptic delays is a potential way to reduce area and allow for more sparse synaptic connectivity on neuromorphic hardware~\cite{goltz2024delgrad}.

The voltage stored in the capacitive neuronal membrane integrates currents originating in ion channels, including ion channels in synapses. Additionally, in biorealistic models, voltage experiences spatial diffusion within the cell itself~\cite{carnevale2006neuron,izhikevich2007dynamical,abi2019arbor}. The synaptic current $i_{syn}$ forming at the postsynaptic neuron can take several forms; still, they all share similar first-order dynamics.
As a prototypical example, we will take a commonly used first-order, direct-synapse model like the ones often used in AI applications. In this model, as shown in equation~\ref{eq:first}, a received spike at time $t_{spk}$ leads to a jump in the synaptic current of the postsynaptic neuron, after which it decreases exponentially with time, with a time-constant of $\tau_{syn}$:

\begin{eqnarray}
    \dot {i_{syn}} = -\frac{i_{syn}}{\tau_{syn}} + \sum \delta(t - t_{spk}) \label{eq:first}
\end{eqnarray}

In biorealistic brain simulations, a double exponential conductive synapse is often used, as shown in equation~\ref{eq:exp2}. This representation models the opening $A$ and closing $B$ of synaptic ion channels in response to neurotransmitter release in addition to the external electrical potential to which these channels connect more explicitly (equations \ref{eq:seconda} and \ref{eq:secondb}).
Importantly, as there is no immediate jump in current, the gradients are considerably easier to calculate at the postsynaptic side.

\begin{eqnarray}
    {i_{syn}}& = (A - B) \cdot (v^{post} - E_{syn}) \label{eq:exp2} \\
    \dot {A} &= -\frac{A}{\tau_{A}} + \sum \delta(t - t_{spk}) \label{eq:seconda} \\
    \dot {B} &= -\frac{B}{\tau_{B}} + \sum \delta(t - t_{spk}) \label{eq:secondb}
\end{eqnarray}

\subsection{Gradients for spiking neural networks}

In simplified neuron models such as LIF, used in most ML models, the reset mechanism (as a result of spikes) causes discontinuities in state variables.
These discontinuities require special handling during gradient calculation, in the form of either surrogate-gradient or exact methods via custom gradients.
In complex neuron models, in which neurons are described by more complicated ODEs but lack a reset mechanism, a continuous voltage trace can be differentiated without custom gradient mechanisms~\cite{landsmeer2024gradient,deistler2024differentiable}.
Detection of these voltage spikes, which is needed for the communication between neurons, is still a simple thresholding mechanism.
A positive crossing over the spike threshold of the membrane potential triggers outgoing spikes on connected synapses. This detection step does require special care for the correct handling of gradients in the same way that spikes in simplified neuron models do.

\subsubsection{Surrogate gradients}

In general, surrogate gradients refer to the design of custom gradients for non-differentiable functions present in a model.
In simplified neuron models, such a function is often the Heaviside step function, and the corresponding surrogate gradient is the so-called SuperSpike~\cite{neftci2019surrogate}. At each timestep $t$, the occurrence $S_t^{j}$ of a spike $j$ is calculated by applying the Heaviside step function $\tilde H$ to the difference between the membrane voltage $V_t$ and the fixed spike threshold value $v_{th}$, that is: $S_t^{j} = \tilde H(V_t^{j}-v_{th})$, where the SuperSpike surrogate gradient defines a custom gradient for the Heaviside step function. The Heaviside step function and its custom gradient, the SuperSpike function, are shown in~\cref{eq:heaviside} and~\cref{eq:superspike}, respectively.
In the case of biophysically realistic neurons (such as Hodgkin-Huxley), a similar approach could be used for the mentioned spike threshold detection. As stated earlier, the dynamics of the state variables do not require such measures as they are differentiable functions.

\begin{align}
x=V_t^{j}-v_{th}&;
    \\
    \tilde H(x) &= 
    \begin{cases}
        1 & x \ge 0 \\
        0 & x < 0 \\
    \end{cases}\label{eq:heaviside}
    \\
    \frac{\partial \tilde H(x)}
         {\partial x} &= \frac{1}{\left(\left|x\right|+1\right)^2}
         \label{eq:superspike}
    \quad
\end{align}

When it comes to performance, surrogate gradients will turn the gradient of the spike events into continuous-in-time, non-zero traces.
This means that a delay-mechanism implementation must not only delay the temporally sparse spike events, but also the dense-in-time surrogate gradient. Because the latter is a dense signal, only an approach which keeps all values in transit in memory can be used for delay handling. An example of this is the ring-buffer used by most ML library-based approaches, which will limit the performance of this method, prompting researchers to explore more efficient gradient-calculation approaches. Such an alternative method must retain the temporal sparsity of the spikes when it comes to their gradient signals.

\subsubsection{Exact gradients}
\label{sec:exactgrads}

\begin{table*}[t!]
    \begin{center}
        \begin{small}
            \caption{
            Overview of existing brain simulators.
            }\label{tbl:related}
            \begin{tabularx}{\textwidth}{l|l|cc|ccc}
                \hline
                & Name & Year~\parnote{Year of first release or publication.}
                & Language/Lib\parnote{TF:~Python+TensorFlow, Torch:~Python+PyTorch, JAX:~Python+JAX} & Gradients\parnote{S:~Surrogate gradients. E:~Exact (event based) gradients. E~+~D:~Exact gradients, including gradients towards spike delays} & Synaptic-Delay Impl. & Biophys. neurons \parnote{Biophysical channels and spatial multicompartmental neurons without restrictions} \\
                \hline
                \parbox[t]{2mm}{\multirow{6}{*}{\rotatebox[origin=c]{90}{\textbf{Traditional}}}}
                & NEURON \cite{carnevale2006neuron}& $<$1998 & C++ & \xmark & P-Queue\parnote{Splay tree priority queue} & \cmark \\
                & NEST \cite{diesmann2001nest} & 2001 & C++ & \xmark & Ring & \xmark \parnote{Minimal support for compartmental neurons is provides in NEST via \texttt{cm\_default}}  \\
                & Brian2 \cite{stimberg2013brian} & 2013 & Python & \xmark & Ring\parnote{Brian has different implementations for homogeneous and heterogeneous queues. Both rely on dynamic memory resizing.} & \cmark \\
                & GeNN \cite{yavuz2016genn} & 2016 & CUDA & \xmark & Ring & \xmark \\
                & Arbor \cite{abi2019arbor} & 2019 & CUDA\parnote{Arbor uses C++ for CPU backend and queue implementation, CUDA on NVIDIA GPUs and translates CUDA to HIP on AMD GPUs} & \xmark & P-Queue\parnote{Tournament tree} & \cmark\\
                & EDEN \cite{panagiotou2022eden} & 2022 & C++ & \xmark & Single-spike & \cmark \\
                & SparseProp \cite{engelken2023sparseprop} & 2023 & Julia & E & P-Queue & \xmark \\
                & GeNN-EventProp \cite{nowotny2025loss} & 2025 & CUDA & E & Ring & \xmark \\
                \hline
                \parbox[t]{2mm}{\multirow{16}{*}{\rotatebox[origin=c]{90}{\textbf{ML-library based}}}}  
                & BINDSnet \cite{bindsnet} & 2018 & Torch & S & \xmark & \xmark \\
                & spytorch \cite{spytorch19} & 2019 & Torch & S & \xmark & \xmark \\
                & NengoDL \cite{rasmussen2019nengodl} & 2019 & TF & S & \xmark & \xmark \\
                & Rockpool \cite{muir_dylan_2019_4639684} & 2019 & JAX & S & \xmark & \xmark \\
                & SpikingJelly \cite{SpikingJelly} & 2020 & Torch & S & Ring & \xmark \\
                & Norse \cite{norse2021} & 2021 & Torch & S & \xmark & \xmark \\
                & snnTorch \cite{eshraghian2021training} & 2021 & Torch & S & \xmark & \xmark \\
                & mlGeNN \cite{turner2022mlgenn,meszaros2025efficient} & 2022 & TF\parnote{mlGeNN is TensorFlow layer on top of GeNN} & E + D & Ring & \xmark \\
                & BrainPy \cite{wang2023differentiable} & 2023 & JAX & S & Ring & \cmark \\
                & Spyx \cite{heckel2024spyx} & 2023 & JAX & S & \xmark & \xmark \\
                & jaxsnn \cite{muller2024jaxsnn} & 2024 & JAX & E & \xmark & \xmark \\
                & jaxley \cite{deistler2024differentiable} & 2024 & JAX & E & \xmark & \cmark \\
                & DelGrad~\cite{golosio2021fast} & 2025 & Torch & E + D & SingleSpike\parnote{In DelGrad, each neuron can only spike once} & \xmark \\
                \hline
            \end{tabularx}
        \end{small}
    \end{center}
    \vspace{-1.5em}
    \parnotes
\end{table*}

As an alternative to the surrogate-gradient approach, exact-gradient calculation has been shown to be possible for LIF cells, by careful treatment of the discontinuous jumps in the membrane voltage~\cite{wunderlich2021event}.
An example of such an exact-gradient approach, EventProp~\cite{wunderlich2021event}, defines the dynamics of the parameter-state Jacobian of each neuron under the presence of spikes (listed in~\cref{sec:ep}). Then, it \textit{manually} derives the adjoint dynamics in the case of LIF neurons for efficient backpropagation.
These are then solved via a forward-in-time pass to obtain spike times and then, a backward-in-time pass with just the spike time information to calculate the needed gradients.

As such, exact-gradient methods show that by careful treatment of spike gradients, temporal sparsity can be retained during gradient calculation~\cite{goltz2024delgrad,wunderlich2021event,goltz2021fast}.
Besides, the DelGrad framework has been proposed which extends exact-gradient calculation by also modeling spike delays, by moving from a voltage-centric view, to a spike-time based communication between layers~\cite{goltz2024delgrad}.
However, there are several limitations associated with these works. All of these methods are constrained to models with simplified LIF cells or their subsets. Some are also limited to certain values of LIF-model parameters~\cite{goltz2024delgrad, goltz2021fast}.
They also require an unconventional approach to simulation, unlike typical setups, as in surrogate gradients, where the user defines the forward dynamics and the automatic gradient (AD) mechanism computes the backward gradients automatically~\cite{wunderlich2021event}.

In conclusion, existing approaches are either memory inefficient because of surrogate gradients, or do not fit into existing frameworks, require manual derivations and do not generalize beyond LIF models.
Therefore, we need a different notion of gradients for spikes, so as to transform current approaches into a simple and general method that:
\begin{enumerate}
    \item fits within existing AD frameworks of ML libraries;
    \item makes use of a sparser spike storage; and
    \item extends to more complex neuron models.
\end{enumerate}

\subsection{Existing brain simulators}

Table~\ref{tbl:related} provides an overview of available brain simulators. They can be classified as either traditional or differentiable, ML library-based.
Traditional brain simulators implement spike-event delivery using priority queues or ring-buffers~\cite{carnevale2006neuron,abi2019arbor,diesmann2001nest,stimberg2013brian,yavuz2016genn}.
ML library-based implementations often do not implement spike delays at all, choosing to directly deliver any spike events at the next timestep to the receiving neuron~\cite{norse2021,eshraghian2021training}. In cases that delayed spike delivery is implemented, either ring buffers are used~\cite{wang2023differentiable} or the implementation allows neurons to spike only once~\cite{goltz2024delgrad}.
Additionally, most of these ML library-based implementations use surrogate-gradient approaches.
For LIF cells, an exact solution to neuron dynamics is known: exact gradients via EventProp~\cite{wunderlich2021event}.
This allows for simulators to implement event-based simulations~\cite{goltz2024delgrad,muller2024jaxsnn,engelken2023sparseprop}.
Among these EventProp implementations, SparseProp~\cite{engelken2023sparseprop} uses 
a SNN training method that uses a binary heap for memory-efficient training, which highlights the potential of using different queue structures for training SNNs with delays.

\subsection{ML-library programming model}

Targetting a variety of accelerators while making use of AD to enable generalization of neuron models, requires use of a ML library.
The ML-library programming model consists of describing a compute graph using high-level operations.
Compilation of the graph happens after specializing the graph on exact types and array shapes.
Examples of such frameworks are JAX~\cite{jax2018github}, TorchScript~\cite{paszke2019pytorch}, and TensorFlow JIT~\cite{tensorflow2015-whitepaper}.

As ML-libraries are mainly developed for AI workloads, most of the development effort is focused on vector and matrix operations. More traditional data structures that require dynamic memory (re-)allocation, like trees, are less supported.
Moreover, as JIT compilation specializes on the shape of arrays, a resize of an array would either lead to JIT-recompilation or is not possible at all inside a JIT-compiled function, severely limiting dynamic memory management. As such, queue implementations need careful design not only with constraints from AI accelerators, but also deriving from the ML-libraries themselves.

\subsection{Custom gradients}

In JAX, efficient, automatic gradient calculation happens through appropriately defined pushforward, Jacobian-vector product (JVP), and pullback, vector-Jacobian product (VJP) operators.
This autograd mechanism, by default, is not aware of the implicit derivatives required for exact spike-event gradients.
Thus, we are required to inject these gradients in JAX' autograd mechanism, via \emph{custom gradients} for the data-structure operations through custom JVP definitions, after which JAX will automatically derive the right VJP.

This main task of the custom JVP is to relate how gradients of the function inputs relate to gradients on the function output. This is often computationally more efficient than calculating the full Jacobian of the function, and using that to transform these gradients. In backpropagation, and backpropagation-through-time, which is preferable in the case where the system has less outputs than inputs, a chain of VJPs is used to transform cotangents at the output to cotangents of the parameters. JAX automatically derives the required VJPs from the JVP, via linearization and transposition of the JVP function~\cite{blondel2024elements}. Clearly, the JVP must be linear in the input tangents.

\section{Mathematical analysis}
\label{sec:methods}

To support a wide spectrum of applications --from AI-oriented SNNs to parameter tuning in biophysically detailed brain simulations-- spike-event queues must be designed to robustly accommodate the full diversity of equations and dynamical regimes encountered across these domains, and fit in existing AD framework to (prevent) manual derivations through more complex neuron models.
These spike-event queues store spikes in transit in either weight space or time-space.
However, spike-time gradients can only be represented in the time direction. Thus, we must derive how to transform from the voltage space to the spike-time space on spike detections, and back to synaptic current gradients on spike delivery.

In this section,
first, the implicit derivatives of spike-gradients are recalled and generalized to arbitrary parameters. Then, we combine these equations with delayed spike event defintion, and a generic exponential synaptic current template. These gradients are then written in a way which makes it possible to define custom gradients in the autograd framework. As an small extra result, we also derive the case for continuous event streams.

\subsection{Implicit function theorem \& EventProp}
\label{sec:ep}

We start the derivation of the delay-enabled gradients following the (forward in time) EventProp equations.
In particular, for a threshold-based spike (spike at $v = v_{th}$), we can derive the spike-time gradient using the implicit function theorem~\cite{yang2014proof}, leading to eq. \ref{eq:wellknown}:

\begin{eqnarray}
\label{eq:wellknown}
    \frac{\D t^{spk}}{\D \theta} &=
    -\frac{1}{\dot v^{pre}} \frac{\D v^{pre}}{\D \theta}
\end{eqnarray}

\noindent where $t_{spk}$ is the spike time, $v^{pre}$ is the pre-synaptic neuron membrane voltage at the point when it crosses the threshold voltage, and $\theta$ is the parameter set of the model.
At the post-synaptic neuron, the reception of a spike leads to an instantaneous jump $\Delta x$ in the state variables $x$ from $x^-$ to $x^+$.
In EventProp, these state variables are $i_{syn}$ and $v^{post}$.
The spike leads to an instantaneous jump in $i_{syn}$, and indirectly, an instantaneous jump in $dv^{post}/dt$.
These updates, both direct and indirect, are then incorporated in the gradient as shown in eq. \ref{eq:eventprop}
~\cite{wunderlich2021event}:

\begin{eqnarray}
\label{eq:eventprop}
    \frac{\partial x^+}{\partial \theta}
    =
    \frac{\partial x^-}{\partial \theta}
    +
    \left(
    \frac{\partial x^-}{\partial t}
    -
    \frac{\partial x^+}{\partial t}
    \right)
    \frac{\partial t^{spk}}{\partial \theta}
    +
    \frac{\partial \Delta x}{\partial \theta}
\end{eqnarray}

\subsection{Differentiable event queues}

\begin{figure*}[h!]
    \centering
    \includegraphics[width=\linewidth]{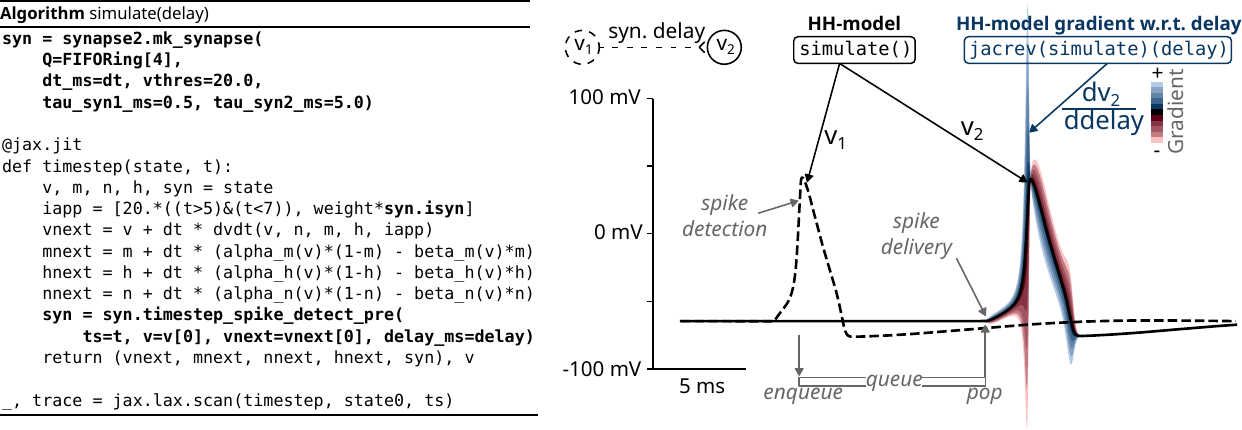}
    \caption{
    ADSEQ works well with existing neuron models and enable easy automatic gradient calculation; here examplified by two 
    Hodgkin-Huxley neurons connected via a double exponential synapse. Synaptic delays are handled by the synapse abstraction introduced in this work, which, when handled by JAX autograd system, will correctly differentiate towards either weights or delays.
    Blue and red areas correspond to the region between $v_2$ and respectively $v_2+0.5\cdot{dv_2}/{d\rm delay}$ and
    $v_2-0.5 \cdot {dv_2}/{d\rm delay}$.
    }
    \label{fig:usage}
\end{figure*}

Reiterating the need for autodifferentiable, sparse and model-independent spike-event queues from~\cref{sec:exactgrads}, we generalize the equations~\cref{eq:wellknown} and~\cref{eq:eventprop} to add the delay gradient and allow arbitrary --not weight only-- parameter-gradients for single and double exponential synapes, and finally show the implementation of these generalized equations in a modern AD framework.

A spike queue takes in a spike delivery time $t^{post}$ (the presynaptic spike $t^{pre}$ delayed by a delay $d$; \cref{eq:mktpost})  and outputs a spike indicator $s(t)$ at the delivery time $t = t^{post}$ (\cref{eq:spkind}) as:

\begin{eqnarray}
   t^{post} &= t^{pre} + d \label{eq:mktpost} \\
   s &= \delta(t - t^{post}) \label{eq:spkind}
\end{eqnarray}

To make these equations fit into an AD framework, we need to find how to carry forward derivatives from the RHS to the LHS of \cref{eq:mktpost}.
If we assume that the spike originates from the crossing of a threshold at $v^{pre}(t^{pre}) = v_{th}$, following equations~\cref{eq:wellknown,eq:eventprop}, and adding an extra gradient towards the delay $d$, we have

\begin{eqnarray}
    \frac{\D t^{pre}}{\D \theta} &=
    -\frac{1}{\dot v^{pre}} \frac{\D v^{pre}}{\D \theta}
    \\
    \frac{\D t^{post}}{\D \theta} &= \frac{\D t^{pre}}{\D \theta} + \frac{\D d}{\D \theta}
    \label{eq:spiketimegrad}
\end{eqnarray}

\noindent At the postsynaptic neuron, the arrival of a spike leads to an instantaneous increase in one or multiple first-order state variables (\cref{eq:first,eq:seconda,eq:secondb}). Using a similar approach to that used in EventProp~\cite{wunderlich2021event}, in general, for any $x\in\left\{i_{syn}, A, B, \ldots\right\}$, with $x^+$ referring to after the increase and $x^-$ to before the increase, we can write:

\begin{align}
    x^+ &= x^- + 1 \qquad \mathrm{at} \;\;  t=t^{post} \label{eq:incr}
    \\
    \dot x &= - \frac{x}{\tau} \label{eq:decr}
\end{align}

\noindent Taking the derivative of \cref{eq:incr} towards parameters $\theta$ and applying the chain rule, we obtain:

\begin{eqnarray}
    \frac{\partial x^+}{\partial \theta}
    =
    \frac{\partial x^-}{\partial \theta}
    +
    \left(
    \dot x^-
    -
    \dot x^+
    \right)
    \frac{\partial t}{\partial \theta}
\end{eqnarray}

\noindent Using \cref{eq:decr} to derive that $\left(\dot x^- - \dot x^+\right) = \frac{1}{\tau}$, we arrive at the final equation for our event queue, at the receiving side:
\begin{eqnarray}
    \frac{\partial x^+}{\partial \theta}
    =
    \frac{\partial x^-}{\partial \theta}
    +
    \frac{1}{\tau}\frac{\partial t^{post}}{\partial \theta}
    \label{eq:receive}
\end{eqnarray}

The \texttt{enqueue()} and \texttt{pop()} functions of our spike-event queues should thus --regardless of how these queue structures store the spikes--
keep track of these gradients in the operations by defining corresponding JVP or VJP operators.

At the level of the neuron, a double-exponential synapse (\cref{eq:exp2}) can be constructed by considering two such variables $x_1$ and $x_2$, each with a separate time-constant, and setting $i_{syn} = \mathrm{const} \cdot(x_2 - x_1)$. This simplifies the implementation, as there is no direct change in neural voltage $v$ or $\dot v$.

\subsection{Continuous signal delays}
Having dealt with discontinuous signals, we will also derive the equations for the case of continuous signals, as for example would be used when delaying a voltage signal.
Here, a buffer can delay a continuous signal $y(t)$ by an amount $d$:

\begin{eqnarray}
   y^{post} &= y^{pre} \left (t - d \right)
\end{eqnarray}

When performing autograd on a time-discretized version of this equation, the gradients will be correct towards $y^{pre}$ but not towards $d$, because $t-d$ will be used in array index operations. That is fine if $d$ is a fixed constant, but if the delay depends on the parameters (for example, when training the delays to find the right value), it is needed to impose the correct gradient, found using the chain rule, to obtain:

\begin{eqnarray}
   \frac{d y^{post}}{dx} &=
   \left. \frac{\partial y^{pre}}{\partial x}
       \right|_{t = t - d}
       +
   \left.
       \frac{\partial y^{pre}}{\partial t}
       \right|_{t = t - d}
       \cdot
       \frac{\partial d}{\partial x}
\end{eqnarray}

When performing time-discretization of such system of equations in an ML library, followed by automatic gradient calculation, the term $\left. \frac{\partial y^{pre}}{\partial x}\right|_{t = t - d}$ is derived automatically, while the term $\left. \frac{\partial y^{pre}}{\partial t}\right|_{t = t - d}\cdot\frac{\partial d}{\partial x}$ has to be implemented with a custom gradient, as AD will not differentiate towards the time argument.

However, since in this case signals are by definition continuous and not sparse in time, no alternative, space-efficient, queue implementation can be used. As such, in this work, we will not further dive into their performance analysis across AI accelerators.

\section{ADSEQ implementation}
\label{sec:impl}

Given the delay-aware spike gradients derived in the previous section, we now need a methodology to \emph{reusably} implement these in a set of event queues for brain simulation.
ADSEQ aim to bring exact delay-gradient to arbitrary brain models by implement event queues for a variety of use cases.
An example is shown in Figure~\ref{fig:usage}. Two Hodgkin-Huxley neurons are simulated using a straightforward translation of the mathematical equations to JAX-expressions. Then, delayed spike event delivery is provided by ADSEQ.
The simulation code does not take further precautions for gradient-calculation, but still is fully differentiable, including towards delays, because the underlying custom gradients.
As described in \cref{sec:bg}, we use JAX as the target framework
due to its general AD primitives and ability to target different accelerators. Our implementation should translate equally to other ML frameworks as well.

To implement ADSEQ, first a method for injecting the derived spike-time gradients and spike-receive gradients into the AD flow is required.
Furthermore, as spike-time gradients existing in the time-direction and not necessarily in the same direction as the queue structures use to represent spikes-in-transit, custom gradients are needed to preserve spike-time gradients in the queue's \texttt{enqueue()} and \texttt{pop()} operations. Finally, different queue datastructures are presented with these principles in mind.

\subsection{Integrating spike-time gradients in autograd}

\begin{lstlisting}[language=Python,caption={Synapse implementation},label={lst:syn},float]
def syn.spike_detect_pre():
    tpost = spike_detect(
      dt_ms, t_ms, vthres, v, vnext, delay_ms)
    queue = jax.lax.cond(tpost != -1,
      lambda: queue.enqueue(
        time_to_timestep_keep_gradient(
          tpost, dt_ms)),
      lambda: queue)
    queue, post_hit = queue.pop(
      time_to_timestep_keep_gradient(
        t_ms, dt_ms))
    isyn = alpha * isyn + \
      receive_spike(post_hit, tau_syn1_ms)
\end{lstlisting}

A custom gradient defined via a custom JVP, in JAX, at the API level, consists of two functions: the regular function, which only operates on primal --normal-- values, and the custom JVP function, which takes in both primals arguments and their tangents --gradients--, and returns a 2-tuple of the regular primal output, and the tangent output -- the tranformed gradient. 

Integrating  the spike-time gradients in autograd thus corresponds to writing this function for the earlier defined  \cref{eq:spiketimegrad} and \cref{eq:receive}.
Spike-detection and -reception logically happen outside the queue, and in ADSEQ, are abstracted away in a single exponential and double exponential synapse class, although the functions shown here can be used standalone. Beyond spike detection and queue handling, the synapse class also handles the dynamics of the synapse as shown in \cref{lst:syn}.

The implementation of spike-time gradient \cref{eq:spiketimegrad} as a JVP is shown in 
as part of the spike-detection function (\cref{lst:detect}).
Spike detection happens via checking whether the current voltage is smaller than the threshold and the next voltage larger than the threshold.
The required $\frac{dv}{dt}$ is estimated from the first forward difference, which when the spike-detection function is used in a forward-Euler simulation, should correspond to the exact voltage-time derivative.
To prevent numerical issues (nans), the estimated $\frac{dv}{dt}$ is set to 1 in the case it is zero. This can never happen during an actual spike detection, thus does not cause further problems.
Finally, \cref{eq:spiketimegrad} is used to calculate the spike-time gradient in the case of spike detection, or 0 is outputted when there is no spike.

The implementation of the spike-receive gradient is shown in  \cref{lst:receive}.
This directly implements \cref{eq:receive} and makes sure that the output gradient is zero when there is no spike.

\begin{lstlisting}[language=Python,caption={Spike-time gradient, \cref{eq:spiketimegrad}},label={lst:detect},float,morekeywords={defjvp}]
@spike_detect.defjvp
def spike_detect_vjp(primals, tangents):
    dt, t, vthres, v, vnext, delay = primals
    _,_,_, v_t, vnext_t, delay_t = tangents
    dvdt = (vnext - v) / dt
    hit = (v < vthres) & (vnext >= vthres)
    dvdt_safe = jnp.where(dvdt==0, 1, dvdt)
    primal_out = jax.lax.select(
      hit, t + delay, -1.)
    tangent_out = jax.lax.select(
        hit, -1/dvdt_safe*v_t + delay_t, 0.)
    return primal_out, tangent_out
\end{lstlisting}
\begin{lstlisting}[language=Python,caption={Spike-receive gradient, \cref{eq:receive}},label={lst:receive},float,morekeywords={defjvp}]
@receive_spike.defjvp
def receive_spike_grad(primals, tangents):
    hit, tau_syn = primals
    tpost_t, tau_syn_t = tangents
    primal_out = jax.lax.select(hit != 0,
        1.0, 0.0)
    tangent_out = jax.lax.select(hit != 0,
        (1/tau_syn * tpost_t), 0.0)
    return primal_out, tangent_out
\end{lstlisting}

\subsection{Preserving spike-time gradients in queue datastructures}

Each queue is implemented as a Python class, that is also JAX PyTree, to store all internal state variables (\cref{lst:class} for an example).
JAX, during the autodifferentiation pass, will assign each state variable a separate primal (\texttt{self}) and tangent (\texttt{self\_t}). We will store the spike time gradients in this tangent queue, which is implicitly generated during AD.
These queues are generic and not tied to brain simulation itself.

The two main functions that operate on the queue are \texttt{enqueue()} and \texttt{pop()}. 
These have their own custom gradients, respectively shown for the SingleSpike example in \cref{lst:enqueue} and \cref{lst:pop}. The enqueue operation in \cref{lst:enqueue} is not different from the default gradient from JAX, but other queue implementations require a custom gradient here to preserve the spike-time gradients.
When we define a custom JVP for the queue \texttt{enqueue()} and \texttt{pop()} operations, one of the inputs and outputs, as this function operates on the queue class itself, is the tangent queue class \texttt{self\_t}. This corresponds to the tangent version of the queue, i.e. the part of the queue that stores spike-time gradients instead of spikes themselves.
As the regular function returns the same output as the primal-to-primal component of the custom JVP, we will only show the custom JVP in the following examples.

\begin{lstlisting}[language=Python,caption={SingleSpikeDrop main class},label={lst:class},float]
import jax
import jax.numpy as jnp
import typing
 
EMPTY = float(0x7fffffff)
 
class SingleSpike(typing.NamedTuple):
    last_spike: jax.Array
    @classmethod
    def init(cls):
        return cls(jnp.array(EMPTY))
    def enqueue(self, n):
        return _enqueue(self, n)
    def pop(self, n):
        return _pop(self, n)
\end{lstlisting}

\begin{lstlisting}[language=Python,caption={Queue enqueue operation},label={lst:enqueue},float,morekeywords={defjvp}]
@_enqueue.defjvp
def _enqueue_grad(primals, tangents):
    self, n = primals
    self_t, n_t = tangents
    primal_out = SingleSpike(jnp.array(n)),
    tangent_out = SingleSpike(jnp.array(n_t))
    return primal_out, tangent_out
\end{lstlisting}
\begin{lstlisting}[language=Python,caption={Queue pop Operation},label={lst:pop},float,morekeywords={defjvp}]
@_pop.defjvp
def _pop_grad(primals, tangents):
    self, n = primals
    self_t, n_t = tangents
    hit = self.last_spike <= n
    primal_out = (jax.lax.cond(hit,
        lambda: SingleSpike(jnp.array(EMPTY)),
        lambda: self),
      hit.astype(self.last_spike.dtype))
    tangent_out = (jax.lax.cond(hit,
        lambda: SingleSpike(jnp.array(0.0)),
        lambda: self_t),
      self_t.last_spike)
    return primal_out, tangent_out
\end{lstlisting}

\subsection{Queue structures}

We implemented our proposed event queues using the data structures presented in this section. 
Spike requests from the queue occur at each timestep; queue event-insertions occur at much lower rates (although they can be quite high for artificial cells).

\begin{itemize}
    \renewcommand\labelitemi{$\triangleright$}
        
    \item \textbf{DoNothing:}\\
    \textit{(no state)}\\
    This implementation drops all inserted spikes and never delivers any spikes. This is added as a reference to measure queue-independent performance and overheads during benchmarking.
        
    \item \textbf{Ring(n):}\\
    \textit{\{buffer: float[delay]\}}\\
    This is a circular delay line which is considered the state of the art in differentiable brain simulators~\cite{wang2023differentiable,SpikingJelly}. This queue implementation is a circular buffer that sums the input spikes, with a size matching the maximum spike delay. A head pointer indicates the beginning of the queue and is advanced at each timestep. Any incoming spike is inserted and summed at location (head pointer + delay). This implementation supports heterogeneous delays up to the buffer capacity, in addition to weighted spikes under the LTI (linear effect) assumption.
    
    \item\textbf{LossyRingDelay(n):}\\
    \textit{\{buffer: float[capacity]\}}\\
    This is a ring implementation with size $n$ smaller than the maximum spike delay, leading to data losses when spike events fall into the same time-bin.
        
    \item \textbf{FIFORing(n):}\\
    \textit{\{buffer: float[capacity], head: int, size: int\}}\\
    This is also a ring implementation with capacity $n$ that queues spike events with First-in, First-out policy. There is a head pointer that advances each time the delivery time of the spike event at the head of the queue the current timestep. This queue implementation drops the incoming spikes when the queue is full. Additionally, it supports multiple spikes at the same time, although it does not perform sorting at insertion thus only supports homogeneous delays.
        
    \item \textbf{SingleSpike(Hold/Drop):}\\
    \textit{\{last\_spike\_time: float\}}\\
    This implementation is a buffer that allows the storage of a single spike. \emph{SingleSpike(Drop)} always replaces the current spike. \emph{SingleSpike(Hold)} does not replace the current spike and instead drops the incoming spike when the queue is full (similar to \emph{FIFORing(1)}). As storage format is not dependent on delays, this implementation does support heterogeneous delays.
        
    \item \textbf{SortedArray:}\\
    \textit{\{buffer: float[capacity]\}}\\
    This implementation has a fixed capacity buffer and a length, and inserts new spike events at the end of the array, then performs a sort operation so the spike events are in order. When delivering a spike, a \texttt{pop} operation is performed and the rest of the array are shifted. In a way, this corresponds to a fixed capacity sorted stack. This queue implementation does support heterogeneous delays, multiple spikes per timestep and arbitrary data storage. The decision of sorting algorithm at spike insertion is left to the ML library implementation of the queue.
    
    \item \textbf{BitArray32:}\\
    \textit{\{buffer: int32\}}\\
    This queue is implemented using a single 32-bit integer. The integer is shifted at each timestep, and if there is a new spike, the most significant bit of the integer will be set to 1. Due to its simplistic, bit-level, design, this queue does not support multiple spikes per timestep or heterogeneous delays. Additionally, because of current ML libraries enforce similarity between primal and tangent data-type representation, it cannot propagate exact (floating point valued) gradients as these would have to be stored as single bits as well.
    
    \item \textbf{BinaryHeap:}\\
    \textit{\{size: int, buffer: float[capacity]\}}\\
    This is a classical heap-queue. Variants of this queue are used in more traditional brain simulators. It is implemented using conditional while-loops, which should allow for some optimization on \textit{von Neumann} platforms but less on dataflow-style architectures.
    
    \item \textbf{BGPQ(1):}\\
    \textit{\{size: int, key\_store: float[8]\}}\\
    This is a heap-based queue with group size of one designed specifically for GPUs~\cite{chen2021bgpq}. It is not expected for this implementation to perform very well as the algorithm achieves parallelism through the group size. However, this queue does support heterogeneous delays, multiple spikes per timestep and arbitrary data storage.
\end{itemize}

\subsection{VJP correctness}

In this work, spike-time gradients are injected via custom JVPs, with JAX' autograd mechanism responsible for generating the VJPs for the backward pass. Correctness of this transform, as well as correctness of the other queue operations, was checked using a set of \texttt{pytest} tests for each queue implementation. The JVP to VJP correctness check is required because not all JAX operations support this operation (e.g. a while loop instead of a fixed-length for loop).

\section{Evaluation}
\label{sec:eval}

In this section, we present the evaluation approach we undertook to benchmark our proposed event queues. The queue implementations described in \cref{sec:impl} are evaluated using different benchmarks on a number of regular and AI-accelerator platforms.

\subsection{Benchmarks}

In order to test the performance of the described queue implementations on different AI hardware, we designed different benchmarks that are presented in this section.

\begin{itemize}
    \renewcommand\labelitemi{$\diamond$}
    \item \emph{Poisson Single/Batched:}
    This is a minimal example that represents a typical ML usage.
    A simple Poisson process ($\lambda$) feeds into a single queue with delay ($d$) in units of $\Delta t$.
    Only a single spike can happen per timestep.
    In effect, queues can fully utilize LTI and homogeneous optimizations, and do not have to merge multiple spike sources.
    Unless stated otherwise,
    $\lambda$=400$\Delta t$ and delay=80$\Delta t$ are used, which roughly corresponds to a
    a biologically realistic 100Hz firing rate and 2ms delays, at $\Delta t = 0.025$.
    The batched benchmark, processes 1000 queues in parallel by default.
    
    \item \emph{Recurrent SNN: Inference/Forward/Reverse:}
    A simple, fully connected recurrent SNN using LIF cells and first-order synaptic dynamics equation (\ref{eq:first}).
    Cells are all-to-all connected, except for self-loops.
    We benchmark the time required for regular simulation (inference), or training (forward/reverse).
    Forward and Reverse refer to respectively forward or reverse autograd calculated gradients, where
    the gradients are calculated with respect to either the delay or the weight matrix.
    
    \item \emph{YY training}
    Timestepped simulation
    
    \item \emph{Eventbased YY training}
    Eventbased simulation
    
    \item \emph{Drop rates:}
    Some queue implementations consider dropping spikes deliberately when reaching capacity.
    Whereas more traditional CPU- or GPU-based implementations might have dynamical reallocation of queues, general AI accelerators using JIT-compiled ML libraries do not have this capability.
    We benchmark some drop rates for typical settings given different \emph{lossy}-spike implementations.
\end{itemize}

\subsection{Platforms}

The AI-accelerator platforms that we used for benchmarking are shown in~\cref{tab:platforms}. Additionally, we performed our benchmarks on a consumer-level CPU as well since brain simulators traditionally target CPUs.

\begin{table*}[h!]
    \centering
    \setlength{\tabcolsep}{7.5pt}
    \begin{tabular}{lcccccc}
        \hline
        \textbf{Name} & \textbf{Year} & \textbf{Process} & \textbf{Transistors} & \textbf{Memory (GB)} & \textbf{F32 Perf.} \\
        \hline
        Intel i7-1195G7 CPU   & 2022 & 4nm  & 80 B  & 16    &  185 GFLOPs~\cite{intelapp} \\
        Nvidia H100 GPU        & 2022 & 4nm  & 80 B  & 80 (HBM)    &  60 TFLOPS \\
        Google TPU v4   & 2021 & 7nm  & 22 B & 32 (HBM)    & 70 TFLOPS (est) \\
        Groq LPU        & 2020 & 14nm & 27 B & 0.23 (SRAM) & 750 TFLOPS  \\
        \hline
    \end{tabular}
    \caption{Evaluation platforms}
    \label{tab:platforms}
\end{table*}

\begin{itemize}
    \renewcommand\labelitemi{\rotatebox{45}{$\diamond$}}

    \item \emph{GPU:}
    The NVIDIA H100 GPU is a general purpose GPUs optimized for AI model execution.
    Among others, it contains 144 Streaming Multiprocessors, each containing 128 FP32 CUDA-cores and 4 Tensor Cores. Tensor Cores provide fast multiply-accumulate operations for TF32 (or lower) precision.
    
    \item \emph{TPU:}
    The TPU v4 is a Tensor Processing Unit developed by Google.
    Its design focuses on dense computation via Tensor Cores (95\% of the area), and sparse computation via Sparse Cores (5\% of the area).
    Each TPU v4 chip contains two TensorCores, with each TensorCore containing four matrix-multiply units (MXUs), a vector unit, and a scalar unit. 
    Sparse Cores, attached to HBM DMA channels, handle gather/scatter operations and also contain a sort unit~\cite{jouppi2023tpu}.
    In previous work, the TPU was very promising with regards to sparse communication, potentially aided by the inclusion of these sparse cores in the fabric~\cite{landsmeer2024tricking}.
    
    \item \emph{LPU:}
    The GroqChip Language Processing Unit (GroqChip LPU) is a deterministic tensor streaming processor, resembling a modified systolic array. The cores are arranged in a grid with horizontal \emph{lanes} and vertical \emph{functional slices}.
    Data moves horizontally, 320 bytes wide per lane, while instructions flow upward vertically. There is no control flow, and all instructions, including data-moves, are scheduled by the compiler before execution, ensuring fully deterministic program execution.
    A functional slice consists of either vector processors, matrix execution modules, switch execution modules (which provide vertical data transfer), or memory modules.~\cite{abts2020think,abts2022groq}.
    
\end{itemize}

\subsection{Experimental setup}

We implemented the queues and their gradients in JAX 0.6.0. Additionally, we used CUDA 12 to deploy our benchmarks on the GPU and Groq-compiler 0.11.0 to run the benchmarks on Groq LPU. Note that this is the latest public release of the Groq compiler and behind newer developments. For all the platforms and benchmarks, correctness of the implementations was verified using randomized pytest-based testing

\section{Results} \label{sec:res}

\subsection{Simulation performance}

\begin{table*}[t]
    \begin{center}
    \begin{small}
        \setlength{\tabcolsep}{4pt}
        \begin{tabularx}{\textwidth}{l|rrrr|rrrr|rrrr}
\hline
\multicolumn{13}{c}{\textbf{Inference}}
\\
\hline
 & \multicolumn{4}{c|}{Poisson (single)} & \multicolumn{4}{c|}{Poisson (batched 10k)} & \multicolumn{4}{c}{R-SNN (10k)} \\
 & CPU & GPU & TPU & Groq & CPU & GPU & TPU & Groq & CPU & GPU & TPU & Groq \\
\hline
DoNothing    & 0.00 & 0.0  & 0.0 & 0.4 &    0.0 & 0.0 & 0.0 & 13.7 &   11.1 &  8.7 & 7.8 &  \\
Ring         & 0.02 & 23.7 & 5.6 & 1.8 &   822.8 & 25.4& 98.9 &  &     6506.0 & 37.9 & 193.1 &  \\
\hline
BitArray32   & 0.00 & 17.6 & 2.7 & &       144.5 & 9.4 & 11.2 &  & & &  &  \\
SingleSpikeD & 0.00 & 25.1 & 2.7 & 1.0 &   151.7 & 9.3 & 11.2 & 15.2 & 15.4 & 10.9 & 11.3 &  \\
SingleSpikeH & 0.00 & 25.1 & 2.7 & 0.7 &   148.0 & 9.2 & 11.2 &  & 51.1 & 10.9 & 11.3 &  \\
\hline
LossyRing[4] & 0.01 & 20.4 & 6.1 & 1.2 &   246.7 & 13.6 & 105.8 &  &  &  &  &  \\
FIFORing[4]  & 0.01 & 26.7 & 5.6 & 3.5 &   260.6 & 14.8 & 193.3 &  & 238.3 & 20.7 & 287.9 &  \\
SortedArray[4]&0.01 & 23.7 & 6.5 & \xmark&1537.7 & 24.4 & 19.8 &  \xmark & 26.8 & 26.8 & 30.4 &  \\
BinaryHeap[7] &0.01 & 63.4 & 8.6 & \xmark&254.7 & 32.8 & 1033.8 &  \xmark & &  &  &  \\
BGPQ1        & 0.02 & 68.8& 32.8 & \xmark&1341.7 & 168.3& 3085 & \xmark &  &  &  &  \\
\hline
\multicolumn{13}{c}{\textbf{Training}} \\
\hline
 & \multicolumn{4}{c|}{R-SNN (forward AD)} & \multicolumn{4}{c|}{R-SNN (reverse AD)} \\
 & CPU & GPU & TPU & Groq & CPU & GPU & TPU & Groq \\
\hline
DoNothing      & 9965    &  217.9 &  M & M & 3.4   & 19.4  & 12.6 &  \\
Ring           & ?       &  M     &  M & M & M     & M     & M  &  \\
\hline
SingleSpikeD   & 83664.0 & 915.3  &  M & M & 83.9  & 30.0  & 40.3 &  \\
\hline
FIFORing[4]    &       ? & 10372.5 & M & M & 670.9 & 61.3 & 88279.4 &  \\
SortedArray[4] & ?       & 10262.5 & M & M & M     & 63.3 & 683.6 &  \\
\hline
\end{tabularx}

    \end{small}
    \caption{
    Inference / simulation.
    Units are microseconds/timestep/neuron (lower is better).
    }
    \label{tab:results}
    \end{center}
\end{table*}

Inference, or forward transient simulation, is how SNNs compute in AI workloads or brain simulations generate their insights for neuroscience.
To benchmark different spike event queue implementation of various hardware platforms, across realistic use cases,
we performed both simple Poisson event-stream experiments and simulated a recurrent spiking neural network.
In general, it is found that CPU performs really well when only a single-queue is simulated (Table~\ref{tab:results}). 
When queue count increases, the AI accelerators perform much better in comparison.
The observed latency in TPU is lower for single queues than for GPU, and is the lowest in the Groq LPU.
This effect is inverted when queue count increases, with the TPU having the most variance in execution time.
At this point, the Groq LPU is unable to fit most implementations on a single chip, due to limitations on code sizes.

\subsection{Training performance}

Training, or parameter tuning, is in general done before putting a model to use for inference. 
This process is much more demanding in terms of computation and memory than inference. In the case of training, compute and memory tradeoff can be made by choosing between forward or reverse automatic gradient methods, which are computation and memory heavy respectively.  An evidence of this is visible in the DoNothing queue case (\cref{tab:results}), where forward AD is much slower than reverse AD, indicating the compute-heavy nature of forward AD. 
A hybrid method, such as checkpointing can be adopted to mitigate the challenges of each of the methods.

\begin{figure*}[t]
    \centering
    \includegraphics[width=1.0\linewidth]{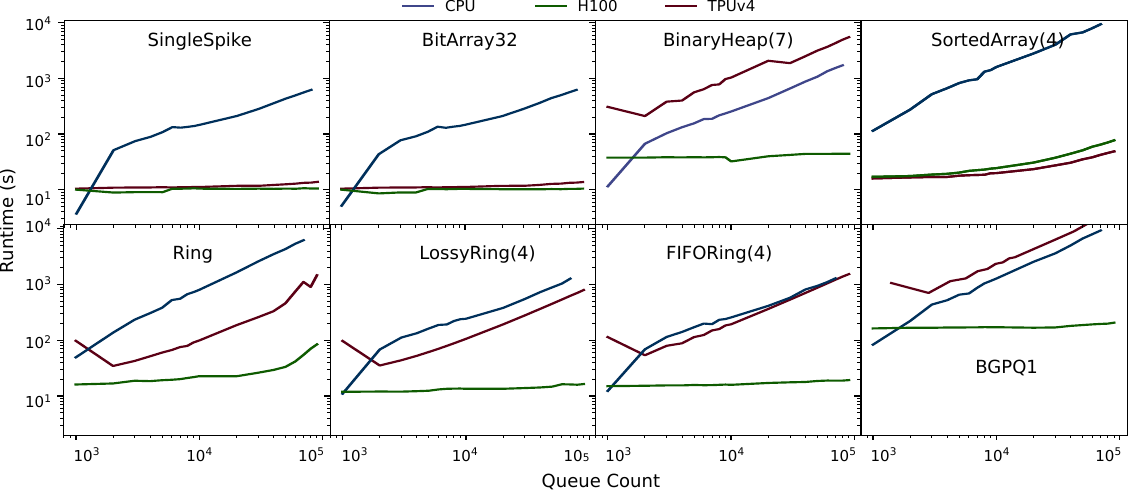}
    \caption{Scaling of event queues with batch size (queue count)}
    \label{fig:sizes}
\end{figure*}

\begin{figure*}[t]
    \centering
    \includegraphics[width=\linewidth]{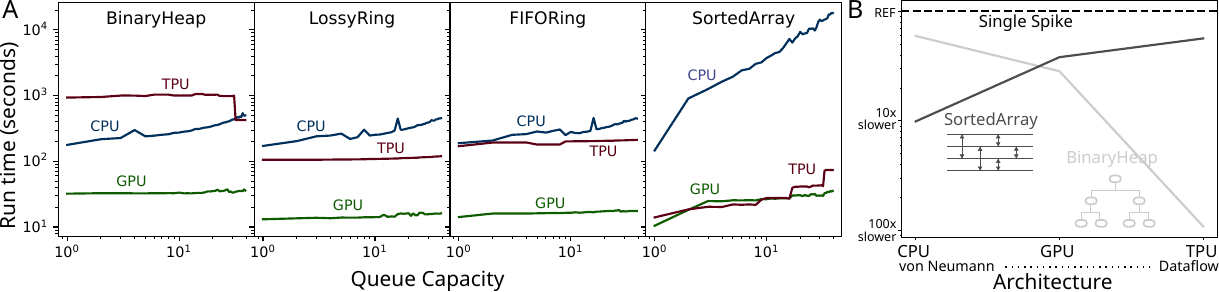}
    \caption{A) Scaling of event queues with queue capacity. B) Architecture Effect}
    \label{fig:caps}
\end{figure*}

\subsection{Scaling queue count}
In general, we find that the CPU performs really well when a small amount of queues are simulated, as shown in~\cref{fig:sizes}. Even compared to other platforms, the CPU outperforms them when it comes to smaller queue count. This could attribute to the lower overhead of execution on CPU compared to other platforms, where the execution overhead is much higher.
When more queues are added to the simulation, CPU performance drops quickly while the performance of GPU and TPU remain more steady.
As can be seen in~\cref{fig:sizes}, a bump around $5\times10^3$ queues is observed across implementations. While subtle, this corresponds to earlier observed artifacts around this number of parallel elements~\cite{landsmeer2024tricking} as well, and thus most likely corresponds to an XLA compilation artifact.

\subsection{Scaling queue capacity}
Queues that support arbitrary delays (BinaryHeap, LossyRing, FIFORing and SortedArray)
have a configurable maximum number of stored future spikes
Increasing the queue's capacity leads to more reliable spike transmission, but also increases memory pressure, decreases memory locality, and in some cases leads to more compute, as can be seen in~\cref{fig:caps}A.
Despite the higher memory demand of increasing queue capacity, in most implementations only a modest increase in runtime with regards to capacity is observed. In the SortedArray implementation however, an increasing capacity leads to poorer performance due to the extra compute required for sorting the larger array.

\subsection{Spike drop rates}

\begin{figure*}[t]
    \centering
    \includegraphics[width=\linewidth]{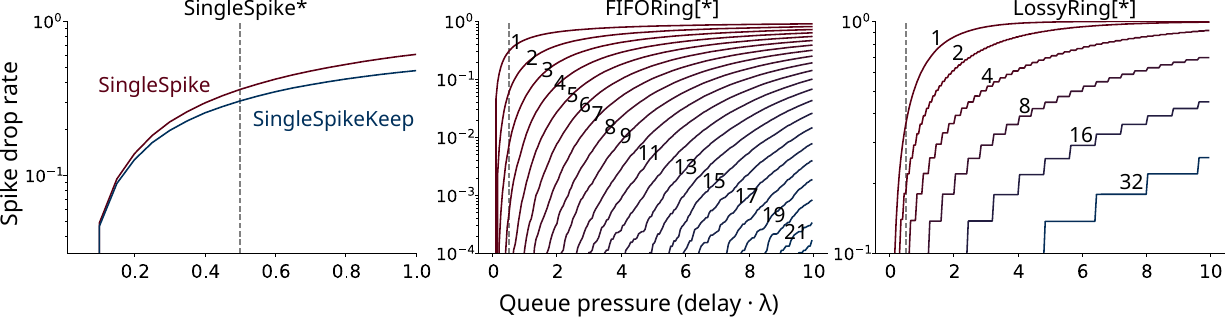}
    \caption{Spike drop rates for different lossy queue implementations given an incoming Poisson process ($\lambda$) and spike delay, both in units of timesteps.}
    \label{fig:droprates}
\end{figure*}

Higher performance is reached for data structures that allow dropping spikes. The number of spikes that are actually dropped depends on the application at hand. The calculated drop rates of different queue implementations for different simulation parameters are shown in~\cref{fig:droprates}. This information would result in better decision making while designing brain simulators.

For a typical brain simulation ($\Delta t = 0.025, f=100Hz, \mathrm{delay}=5ms$), the queue pressure  $\mathrm{delay}/\lambda$ might have a value of 0.5. This suggests that even a very small FIFORing might be enough to have negligible spike drop rates.

\section{Discussion}
\label{sec:discuss}

\subsection{Architecture effect}

The benchmarked hardware architectures range from classical von Neumann (CPU) to deterministic dataflow (Groq LPU).
The GPU and TPU can be considered intermediates between these two paradigms, with the GPU being closer to CPU and the TPU being closer to dataflow in terms of architecture. We expect different queue implementations to vary across this spectrum. To asses this, for each hardware platform, queue performance was normalized against the SingleSpike baseline, as can be seen in~\cref{fig:caps}B.
As expected, binary heaps perform better in more classical architectures. At the same time, the sorting implementation performs better as the underlying hardware moves toward more dataflow-style architectures. 

\subsection{Platform choice}

In general, in high batch counts, memory utilization of the queue seems to be the most important factor for it performance.
In all platforms, when admissible by simulation requirements, SingleSpike or BitArray32 queue implementations perform exceptionally well, showing both the lowest latency and smallest scaling effect. This is not unexpected as these queues can only store a handful of spike events compared to other more sophisticated implementations.
In general, FIFORing should be preferred over Ring on GPUs for large batch sizes, due to its memory size. Furthermore, SortedArray performs very well on TPU, most likely due to the specialized sorting procedures in the sparse cores.
Additionally, CPU offers a clear advance for small simulations, when the entirety of the queues fit in fast local caches.

\subsection{Towards hardware with explicit delay support}

Dataflow oriented hardware architecture are very suited for AI workloads, which mostly consist of predictable and dense vector/matrix operations.
To accommodate for increasing sparse operations in these workloads, including embedding retrieval, a small amount of area dedicated to sparse operations, exemplified by TPU Tensor and Sparse Cores, can lead to a huge leap in performance on sparse-dense workloads~\cite{landsmeer2024tricking}.
However, when delays are included in the workload, this performance benefit thins.
It is shown in similar workloads with delays that explicit delay handling mechanisms, and even exploiting delay property of the application, gives extensive benefits to the performance of the system~\cite{movahedin2025huma}. 
Additionally, advances in materials science might lead to spike-based computing platforms that deviate even further from both von Neumann or dataflow architectures~\cite{landsmeer2025efficient}.

\subsection{Diverging primal and tangent data structures}

JAX, and other ML libraries, are limited in how much the \emph{primal} (regular) and \emph{tangent} (derivative) data structures can diverge.
For example, in JAX, the tangent data type of an integer is a float0 type, thus has no storage in the tangent space, which is useful for efficient gradients for data structures with indexing variables like a FIFORing queue.
On the other hand, a spike counter implementation like Ring is now forced to use float representation in primal space to also store a float for the tangent space.
This also means that a structure like a BitArray32 can not store tangent spikes.

In general, it might be very beneficial to have different data structures in primal and tangent space, something that might be possible in future versions of ML libraries.
For the queue implementations in this work, custom JVPs were defined for any queue which that could support gradients within the JAX framework.
The primal and tangent versions of the queues were implemented using the same data-types.
The obvious solution is to use the same queue implementation for both spikes and their gradients. Ideally, a mixture of implementations can also be used if memory is an issue. For example, one could use bit arrays for the binary encoding of spike existence, and use a lossy storage buffer to approximate the memory heavy gradients.

\subsection{Injecting gradients into traditional brain simulators}

Traditional brain simulators seem to be superseded by ML library-based implementations, as they are not designed for gradient-based tuning.
One exception is GeNN, which has implemented the EventProp algorithm in their CUDA codebase ~\cite{nowotny2025loss}.
Biophysical realistic brain simulators like Neuron, Arbor or EDEN do not support gradient calculation and would require significant development effort to support this from the simulator level. However, we have shown in previous work that gradients can be retrofitted in traditional biorealistic brain simulators, by modification of the neuron models, at the level of single cells~\cite{landsmeer2024gradient}.
In this work, we have shown that traditional event queues can still be used for network-level gradients, if they support sending a float message value with each spike, encoding the spike gradient.
Together, this could pave a way for providing gradients for large-scale morphologically detailed biophysical networks, without dropping existing traditional brain simulators.

\section{Conclusions} \label{sec:conc}

Trainable event delays lead to more efficient encoding in SNNs and increased realism in brain models.
Implementing delay queues, both with and without gradients, allows for considerable freedom in implementation. With AI accelerators offering a range from traditional von Neumann to dataflow-style programming, data structure choice has a large impact on platform-specific performance.
In this work, we derived the equations for generalized queue gradients. Additionally, we investigated how data-structure choice in ML libraries affects performance across different AI accelerators.
It was found that, CPUs, also when targeted from a high level ML library, still prefer tree-based or FIFO structures. GPUs, when memory permits, perform very well with existing ring-buffers. However, in training or larger network sizes, reduced memory pressure from FIFORing and SortedArray leads to the ability to simulate or train much larger networks. On TPUs, the sorted array performs orders of magnitude better than the other implementations, both in inference and training, potentially via the dedicated sorting unit in the sparse cores. In all cases, dropping spikes allows for an easy performance-accuracy tradeoff.
Not limited to time-discretized simulation, ADSEQ implementations should also automatically allow for fully event-based simulation, without changes to the gradient-calculation.
Future work, when allowed for by ML-libraries, could explore how primal and tangent representations could diverge for further performance-accuracy tradeoff.
More broadly, while more traditional data structures like FIFO buffers or trees are often neglected in ML-library based implementations, it seems that in certain cases these might have merit over current dominant simple data-structures.

\subsection*{Acknowledgments}

This paper is partially supported by the European-Union Horizon Europe R\&I program through projects SEPTON (no. 101094901) and SECURED (no. 101095717), and through the NWO - Gravitation Programme DBI2 (no. 024.005.022) and NWO-LSH Program INTENSE (TTW/00798883). 
This work used the Dutch national e-infrastructure with the support of the SURF Cooperative using grant no. EINF-10677. The RTX6000 used for this research was donated by the NVIDIA Corporation. 
This research is supported with Cloud TPUs from Google's TPU Research Cloud (TRC).

\bibliography{main}

\end{document}